\documentclass[letterpaper, 10 pt, conference]{ieeeconf}  

\IEEEoverridecommandlockouts                              

\usepackage{graphics} 
\usepackage{epsfig} 
\usepackage{mathptmx} 
\usepackage{times} 
\usepackage{amsmath} 
\usepackage{amssymb}  
\usepackage{caption}
\usepackage{subcaption}

\usepackage{algorithm}
\usepackage{algpseudocode}
\usepackage[algo2e]{algorithm2e}
\usepackage{booktabs} 
\usepackage{multirow}
\usepackage[table,x11names]{xcolor}
\usepackage{balance}

\newcommand{\cmark}{\textcolor{green!80!black}{\ding{51}}}
\newcommand{\xmark}{\textcolor{red}{\ding{55}}}
\usepackage{pifont}

\usepackage{cite}

\title{\LARGE \bf
Gradient-based Trajectory Optimization with \\ Parallelized Differentiable Traffic Simulation
}

\author{Sanghyun Son*$^{1}$, Laura Zheng*$^{1}$, Brian Clipp$^{2}$, Connor Greenwell$^{2}$, Sujin Philip$^{2}$, and Ming C. Lin$^{1}$
\thanks{*Equal contribution}
\thanks{$^{1}$The authors are with (1) Department of Computer Science, University of Maryland at College Park, MD, U.S.A., (2) Kitware. E-mail: \{shh1295,lyzheng,lin@umd.edu\}, \{brian.clipp,connor.greenwell@kitware.com,\}, \{philip.sujin@gmail.com\}}%
}

\begin{document}

\maketitle
\thispagestyle{empty}
\pagestyle{empty}

\begin{abstract}

We present a {\em parallelized} differentiable traffic simulator based on the Intelligent Driver Model (IDM), a car-following framework that incorporates driver behavior as key variables. Our vehicle simulator efficiently models vehicle motion, generating trajectories that can be supervised to fit real-world data. By leveraging its differentiable nature, IDM parameters are optimized using gradient-based methods. With the capability to simulate up to \textit{2 million} vehicles in real time, the system is scalable for large-scale trajectory optimization. We show that we can use the simulator to filter noise in the input trajectories (\textit{trajectory filtering}), reconstruct dense trajectories from sparse ones (\textit{trajectory reconstruction}), and predict future trajectories (\textit{trajectory prediction}), with all generated trajectories adhering to physical laws. We validate our simulator and algorithm on several datasets including NGSIM and Waymo Open Dataset. The code is publicly available at: https://github.com/SonSang/diffidm.

\end{abstract}

\section{INTRODUCTION}

For design, prototyping, testing, and evaluation of dynamical systems, we often rely on computer simulations that replicate reality, such as urban dynamics or natural phenomena. Traffic simulation, for example, is often used to optimize traffic light policies~\cite{tang2019cityflow, son2022differentiable} or to train autonomous vehicles in varied scenarios~\cite{highway-env, wu2017flow}. Traffic models come in different scales, in this paper we focus on microscopic models that simulate the movement of {\em every individual vehicle}~\cite{gazis1959car, gazis1961nonlinear, newell1961nonlinear, krauss, bando1995dynamical, gipps1981behavioural, jiang2001full, treiber2000congested}, as opposed to macroscopic models that simulate {\em aggregate} traffic flow~\cite{lighthill1955kinematic, richards1956shock, payne1971model, whitham2011linear, aw2000resurrection, zhang2002non}. We adopt the Intelligent Driver Model (IDM)~\cite{treiber2000congested}, an ODE describing car-following behavior, as it is widely used in diverse traffic simulators~\cite{krajzewicz2012recent, dosovitskiy2017carla, highway-env, gulino2024waymax}.

When developing a traffic simulator, we aim for both efficiency and differentiability. The simulator must be efficient to gather enough data to solve the problem without excessive cost. This is especially important as traffic problems continue to grow in scale~\cite{tang2019cityflow, zhang2024moss}. Optimization techniques are also key, and they can be classified as either {\em gradient-based} or {\em gradient-free}. Among them, gradient-free methods typically require significantly more data, because of its inferior sample efficiency. To ensure sample efficiency and reduce costs in training, testing, and evaluation, we designed our simulator to be {\em differentiable} via auto-differentiation frameworks.


\begin{figure}
    \centering
    \begin{subfigure}[b]{0.49\linewidth}
        \centering
        \includegraphics[width=\linewidth]{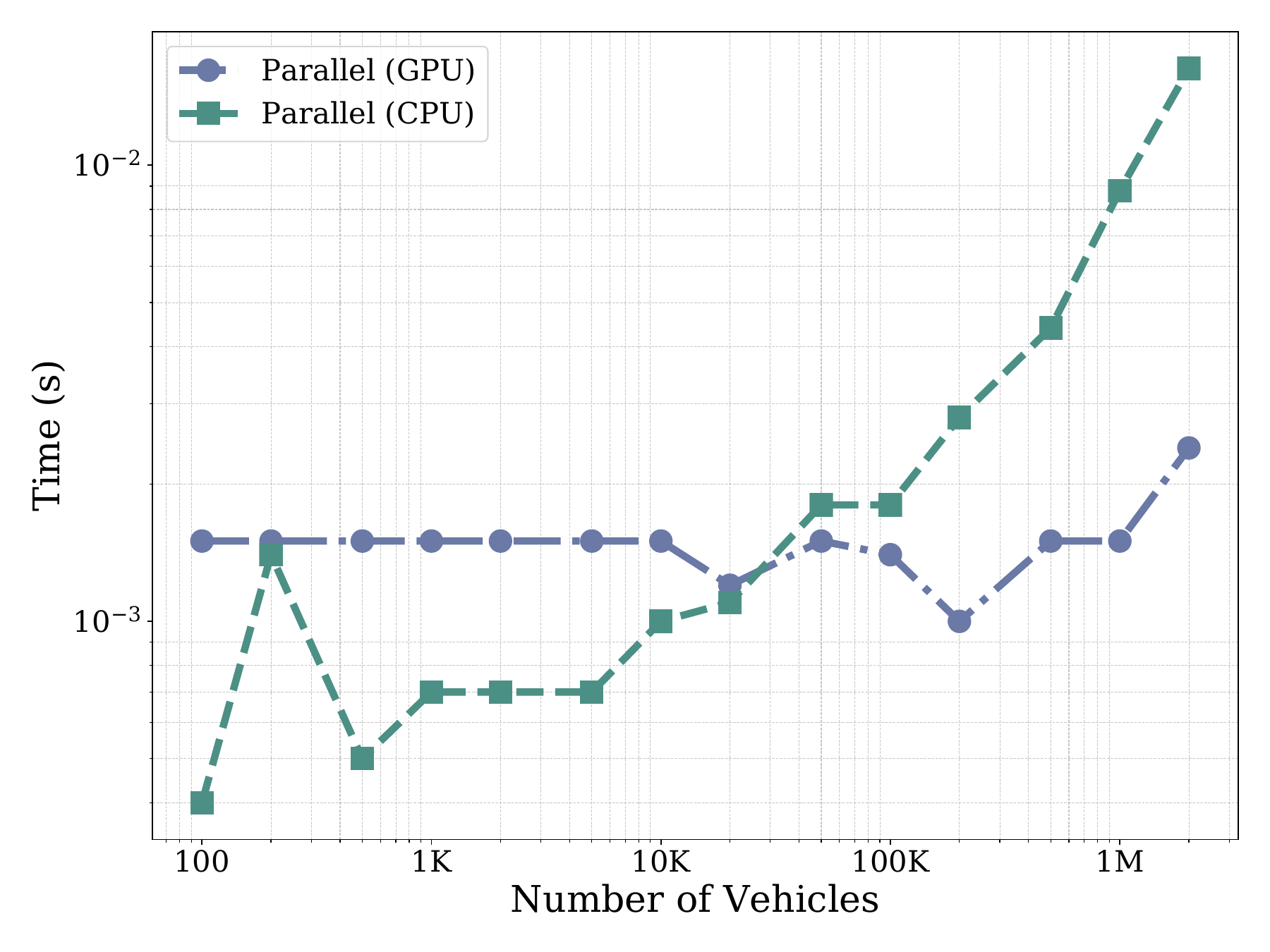}   
        \caption{Forward pass}
    \end{subfigure}
    \begin{subfigure}[b]{0.49\linewidth}
        \centering
        \includegraphics[width=\linewidth]{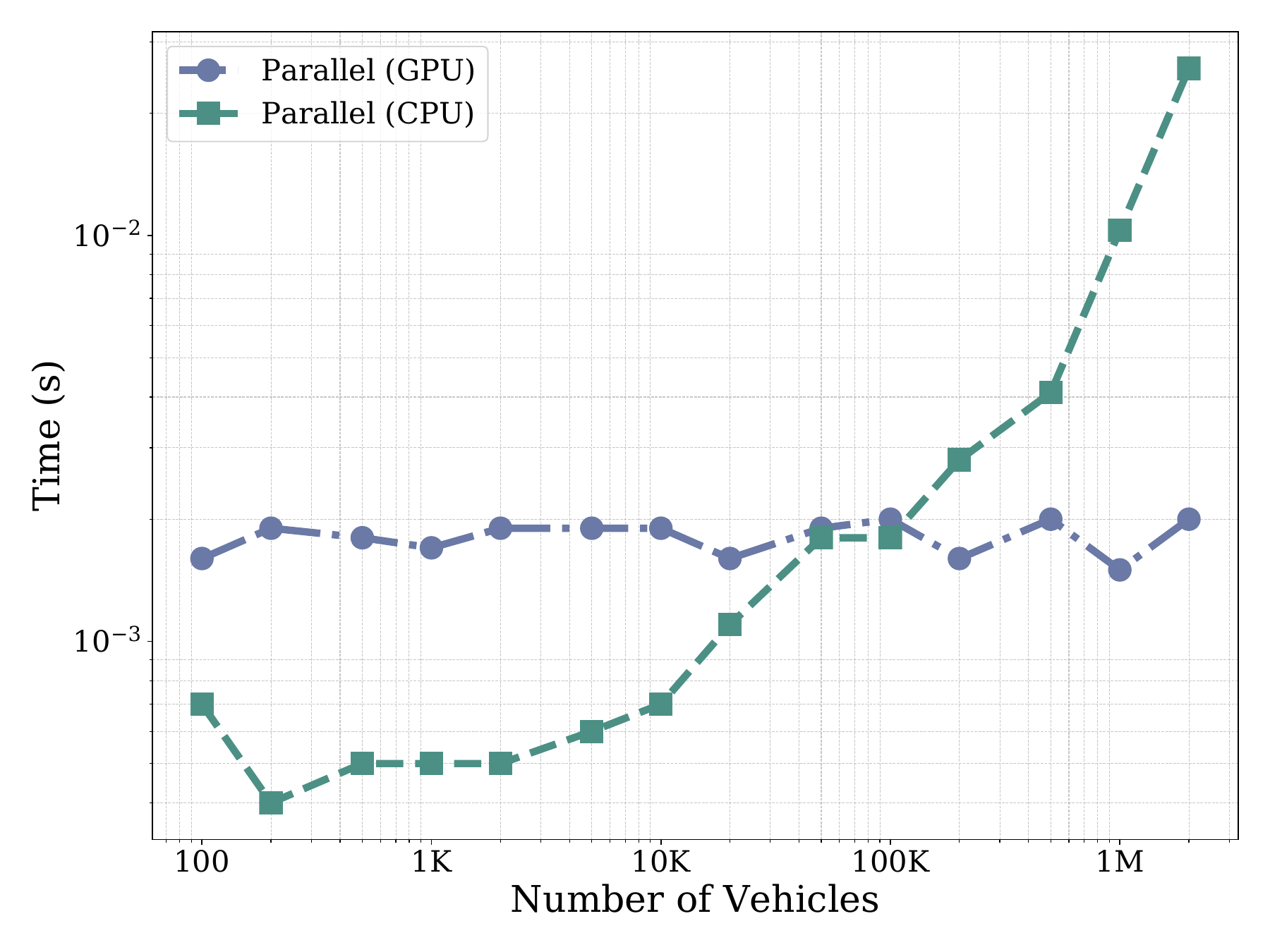}    
        \caption{Backward pass}
    \end{subfigure}
    \caption{\textbf{Computational cost of our traffic simulator.} Using either (multi-threaded) CPU or GPU, our simulation takes less than $30$ milliseconds \textit{per timestep} to process $2$ million vehicles in both (a) forward and (b) backward pass.}
    \label{fig:cost-comp-device}
    \vspace{-2em}
\end{figure}

Differentiable simulation has successfully bridged real-world dynamical systems with the power of deep learning~\cite{de2018end, degrave2019differentiable, qiao2021efficient, freeman2021brax, xu2022accelerated, son2024gradient}. It allows a state at one timestep to be differentiable with respect to that of the previous one, enabling gradient-based optimization techniques that require fewer samples than gradient-free methods. The differentiable programming paradigm for traffic modeling has gained traction recently~\cite{andelfinger2021differentiable, itra2021, son2022differentiable, zheng2023traffic}, including data-driven approaches~\cite{gpudrive_Kazemkhani_Pandya_Cornelisse_Shacklett_Vinitsky_2024}. However, these methods often suffer from computational inefficiencies such as sequential processing and non-differentiable logic, limiting their practicality for real-time autonomous systems.

To address these limitations, we propose a {\em parallelized} differentiable vehicle traffic simulator that achieves both efficiency and differentiability. Our implementation supports both CPU and GPU parallelization and can simulate up to \textit{2 million} vehicles in real-time (Figure~\ref{fig:cost-comp-device}). We also introduce several differentiable modifications to the original IDM to eliminate unrealistic vehicle behaviors (e.g., backward motion), which were not addressed in previous work~\cite{andelfinger2021differentiable, son2022differentiable}. This simulator can handle various trajectory optimization tasks, including filtering, reconstruction, and prediction (Figure~\ref{fig:trajectory-problems}). Specifically, in trajectory prediction, we explicitly incorporate road network and lane membership information, which existing frameworks have not done. Our main contributions are summarized as follows:
\begin{itemize}
    \item We present an efficient traffic simulator based on IDM, which can simulate up to 2 million vehicles in real time using either {\em CPU or GPU parallelization}.
    \item We derive and implement a {\em differentiable IDM} layer that is guaranteed to generate realistic vehicle motions. Using the gradients from this layer, we can leverage sample-efficient gradient-based optimization schemes for traffic optimization problems; this greatly enhances the feasibility of traffic simulation in end-to-end deep learning.
    \item We validate our method by solving {\em large-scale trajectory filtering, reconstruction, and prediction} for autonomous driving on various datasets. Our trajectories are guaranteed to follow the physical laws, while the existing methods may not always do.
\end{itemize}

\section{Related Work}

\subsection{Traffic Simulators}

Most microscopic traffic simulators are based on car-following models, which describe vehicle acceleration using driver parameters like comfortable acceleration. Notable examples include the Newell Model~\cite{newell1961nonlinear}, the Gipps Model~\cite{gipps1981behavioural}, the Krauss Model~\cite{krauss}, and the Intelligent Driver Model (IDM)~\cite{treiber2000congested}. Popular simulators such as SUMO\cite{krajzewicz2012recent}, HighwayEnv~\cite{highway-env}, CARLA~\cite{dosovitskiy2017carla}, and FLOW~\cite{wu2017flow} use IDM to model vehicle movements, while others like MATSIM~\cite{matsim} and VISSIM~\cite{vissim} employ similar car-following models. IDM's widespread use in simulators influenced our choice. However, these IDM-based simulators lack parallelization and differentiability, making them less suited for solving large-scale traffic problems.

Recent advancements in machine learning and GPU hardware have led to the development of parallelized or differentiable traffic simulators for large-scale traffic simulation and urban mobility optimization. MOSS~\cite{zhang2024moss} uses GPU power for city-scale traffic simulation based on IDM and MOBIL~\cite{mobil_Treiber_Kesting_2009}, but it lacks differentiability, preventing application with gradient-based optimization and deep learning. Other approaches have applied differentiable methods to traffic optimization and control~\cite{andelfinger2021differentiable, son2022differentiable, itra2021}, but they are not as efficient as MOSS. Data-driven simulators like Waymax~\cite{gulino2024waymax} and GPUDrive~\cite{gpudrive_Kazemkhani_Pandya_Cornelisse_Shacklett_Vinitsky_2024} offer both GPU acceleration and differentiation but focus on replaying real-world trajectory logs from the Waymo Open Dataset (WOMD)~\cite{womd_Ettinger_2021_ICCV}, without reactive traffic behaviors. To the best of our knowledge, our simulator is the {\em first model-based traffic simulator that achieves both high efficiency and differentiability for optimization, learning, and control} (Table~\ref{tab:appendix_simulator_comparison}).

\subsection{Trajectory Optimization}

\begin{figure}
    \centering
    \includegraphics[width=0.8\linewidth]{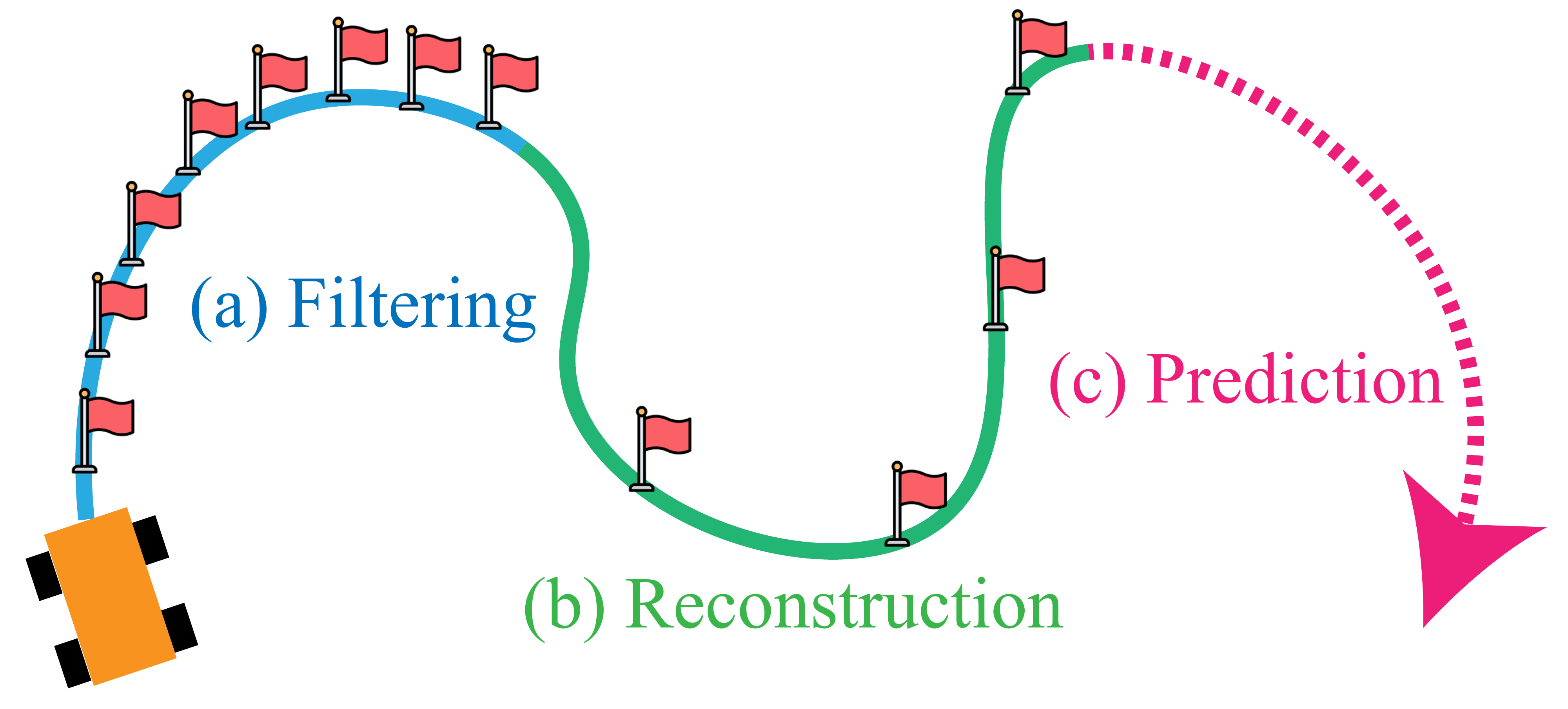}
    \caption{\textbf{Trajectory Optimization Problems.} For the given data points (red flags), we can fit a simulated trajectory to them by optimizing IDM variables. (a) When data points are dense, we can filter physically inaccurate noises in the original trajectory. (b) When they are sparse, we can reconstruct dense trajectories. (c) We can even predict future trajectories based on the IDM variables.}
    \label{fig:trajectory-problems}
    \vspace{-1em}
\end{figure}

We address three trajectory optimization problems: trajectory filtering, reconstruction, and prediction (Figure~\ref{fig:trajectory-problems}). For trajectory filtering, we assume we have dense observations (e.g., vehicle positions) that can be simply concatenated to construct a dense trajectory. However, numerical derivation on this trajectory amplifies noise, leading to physically impossible estimates of speed or acceleration~\cite{fard2017new} (Figure~\ref{fig:traj-profile}). To address this, various filtering methods have been proposed, including moving averages~\cite{duret2008estimating, ossen2008validity, thiemann2008estimating}, spline smoothing~\cite{jun2006smoothing}, and wavelet analysis~\cite{fard2017new}. While this has been studied under ``trajectory reconstruction," we use the term ``trajectory filtering" to differentiate it from our reconstruction task, which focuses on generating dense trajectories from sparse data points -- essentially an interpolation task. Unlike previous model-free approaches, {\em our model-based approach is guaranteed to generate physically plausible trajectories}.

In the trajectory prediction task, many data-driven methods forecast future trajectories based on historical data and road context~\cite{seff2023motionlm, nayakanti_wayformer, zhou2023query_qcnet, shi2023mtr, shi2022motion, mppp_multipath_varadarajan_2022}. Current state-of-the-art methods use transformer architectures to model trajectories as sequences of positions, similar to natural language processing. However, none incorporate explicit micro-simulated dynamic states into the modeling process, instead relying on highly parameterized models to implicitly capture traffic dynamics during training. This gap exists due to the lack of fast, differentiable traffic simulators—most are either sequential or non-differentiable. While some differentiable environments are used in reinforcement learning tasks~\cite{son2024gradient, son2022differentiable, qiao2020Scalable}, no such simulators are available for imitative or open-loop methods.

\section{Method}

\begin{figure*}
    \centering
    \includegraphics[width=\textwidth]{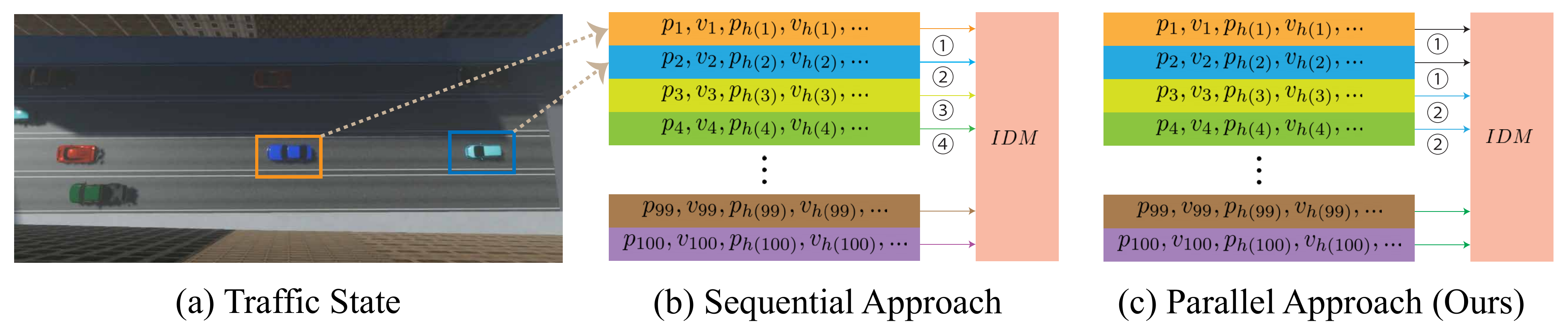}
    \caption{\textbf{Overall Framework.} In our traffic simulator, for each frame, (a) we first collect variables for each vehicle in the scene to use in IDM. These variables include the position ($p_i$) and velocity ($v_i$) of each vehicle, and its relationship to its leading vehicle ($p_{h(i)}, v_{h(i)}$). Since we can apply IDM to each vehicle's variables independently, we can process them in parallel, rather than sequentially. In (b) and (c), the process ordering is depicted with colors and numbers. In (c), we assume there are two computational units that can run in parallel.}
    \label{fig:parallel-overview}
    \vspace{-1em}
\end{figure*}


\subsection{Preliminary}
\label{sec:prelim}

 Our traffic simulator is built on the Intelligent Driver Model (IDM)~\cite{treiber2000congested}, which assumes that a vehicle's acceleration is determined solely by its relationship to the vehicle directly ahead in the same lane. Specifically, if the position and speed of the $i$-th vehicle at time $t$ are denoted as $p_{i}(t)$ and $v_{i}(t)$, and the index of its leading vehicle is $h(i)$, we compute $\Delta p$ (distance gap) and $\Delta v$ (speed difference) as follows:
\begin{align*}
    {\Delta p} &= p_{h(i)}(t) - p_{i}(t) - {length}_{h(i)} \\
    {\Delta v} &= v_{i}(t) - v_{h(i)}(t).
\end{align*}

Based on these variables, we can compute the optimal spacing $s_{\mathrm{opt}}$ and acceleration $a_{i}(t)$ of the $i$-th vehicle as
\begin{align}
    s_{\mathrm{opt}} &= s_{\min} + v_{i}(t)T_{\mathrm{pref}} + \frac{v_{i}(t){\Delta v}}{2\sqrt{a_{\max}a_{\mathrm{pref}}}} \\
    a_i(t) &= a_{\max}\left[1 - {\left(\frac{v_i(t)}{v_{\mathrm{targ}}}\right)}^{\delta} - {\left(\frac{s_{\mathrm{opt}}}{\Delta p}\right)}^{2}\right].
\label{eq:idm}
\end{align}

The formulation for IDM above includes several hyperparameters that describe the vehicle's behavioral traits: $a_{max}$ and $a_{pref}$ for maximum and preferred acceleration, $s_{min}$ for the minimum distance gap, $T_{pref}$ for the preferred time to maintain current speed, and $v_{targ}$ for the target speed. In this paper, we use meters (m) for position, meters per second (m/s) for speed, and meters per second squared (m/s²) for acceleration.

After we obtain the acceleration of each vehicle at time $t$, we compute its position and speed at the next time step, $t + \Delta t$ using Euler integration as follows:
\begin{align}
\begin{split}
    p_{i}(t + \Delta t) = p_{i}(t) + \Delta t \cdot v_{i}(t), \\
    v_{i}(t + \Delta t) = v_{i}(t) + \Delta t \cdot a_{i}(t).
\end{split}
\label{eq:euler}
\end{align}

The time step $\Delta t$ can be adjusted depending on the scenario. While a differentiable version of this formulation has been proposed in several prior works~\cite{son2022differentiable, andelfinger2021differentiable}, these implementations often lack parallelization and result in unrealistic behaviors, such as negative speed. In the following sections, we demonstrate how to address these limitations.

\subsection{Parallelization Scheme}
\label{sec:parallel}
Figure~\ref{fig:parallel-overview} illustrates the overall framework of our simulator. We first collect the necessary per-vehicle information for each simulation frame to compute acceleration using IDM. As shown in Figure~\ref{fig:parallel-overview}(b) and (c), each vehicle's data can be represented as independent data blocks. Due to this independence, instead of processing the blocks sequentially (Figure~\ref{fig:parallel-overview}(b)), we process them in parallel (Figure~\ref{fig:parallel-overview}(c)). Additionally, the IDM kernel is implemented in a differentiable manner. With this parallelization, our simulator can simulate up to 2 million vehicles in real-time on either CPU or GPU and efficiently compute gradients (Figure~\ref{fig:cost-comp-device}).


\subsection{Unrealistic Behaviors}
\label{sec:idm-modify}

We observed that IDM often estimates negative accelerations that are physically impossible or end up with negative speed. To prevent these cases, we first set the lower bound of $s_{opt}$ to 0, as it should be a positive value by its conceptual definition (`optimal spacing'). Then, we set the lower bound of $a_{i}(t)$ to $a_{lb} = \max(-v_{i}(t) / \Delta t, a_{min})$, where $a_{min}$ is a new hyperparameter that corresponds to the (physically valid) maximum deceleration. Note that we can guarantee that the next speed estimated from this acceleration term is non-negative, as we use Euler integration (Eq.~\ref{eq:euler}).
We choose to use Euler integration, as opposed to other time integration methods such as Runge-Kutta, for its simplicity. 
While other methods may be more numerically stable, we find that Euler works well with a sufficiently small timestep $\delta t$ and minimal computing cost.

To implement these lower-bound operations in a differentiable way, we use the softplus function as follows:
\begin{align*}
    s_{opt}^{*} &= \log{(1 + \exp{(s_{opt})})} \\
    a_{i}(t)^{*} &= a_{lb} + \log{(1 + \exp{(a_{i}(t) - a_{lb})})}.
\end{align*}

Note that $a_{i}(t)^{*}$ is guaranteed to exist in $[a_{min}, a_{max}]$.

\section{Application}
\label{sec:application}

In this section, we provide our solutions to the trajectory optimization problems using our differentiable simulator.

\begin{figure*}
    \centering
    \begin{subfigure}[b]{0.18\linewidth}
        \centering
        \includegraphics[width=\linewidth]{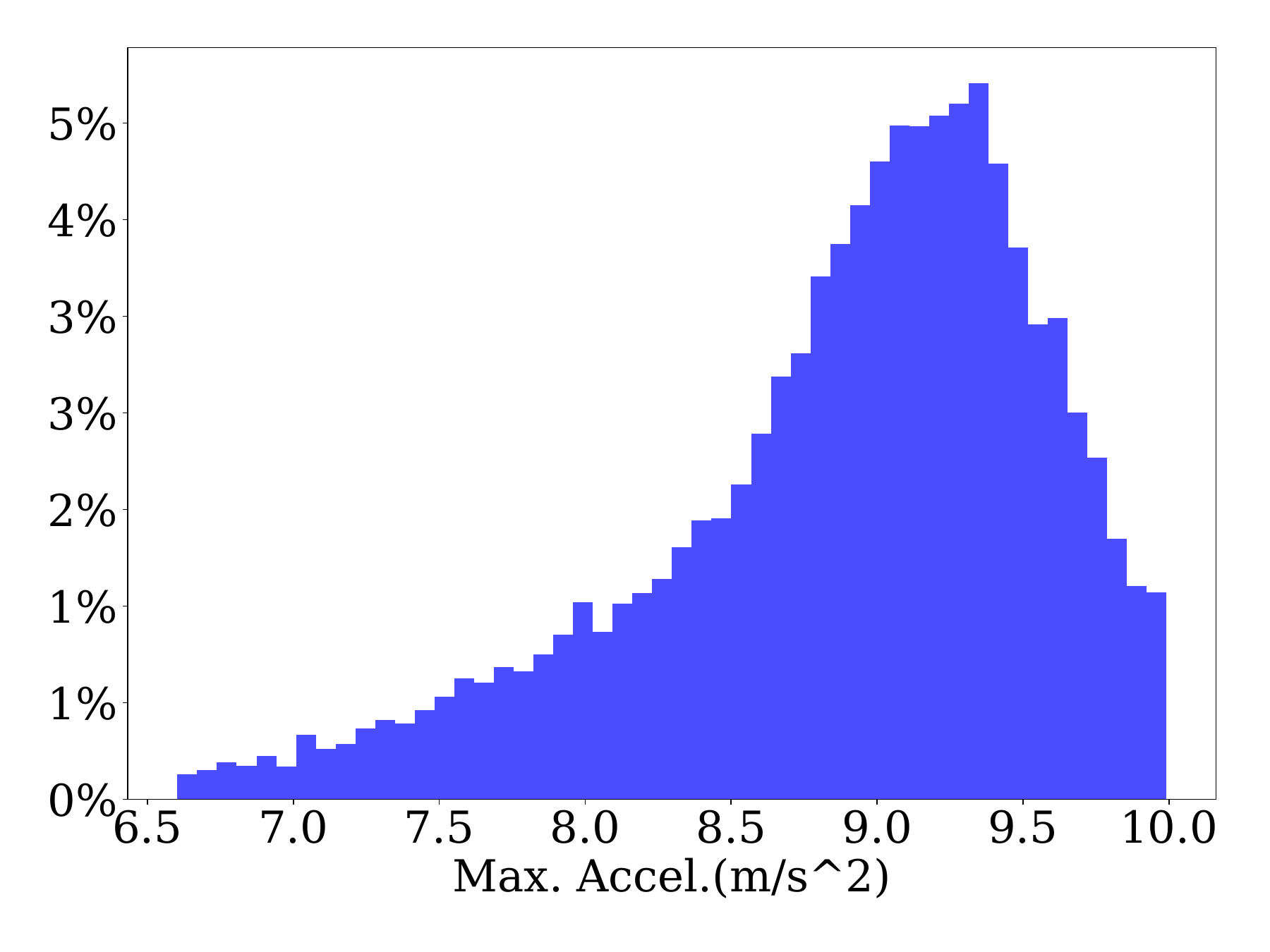}
    \end{subfigure}
    \begin{subfigure}[b]{0.18\linewidth}
        \centering
        \includegraphics[width=\linewidth]{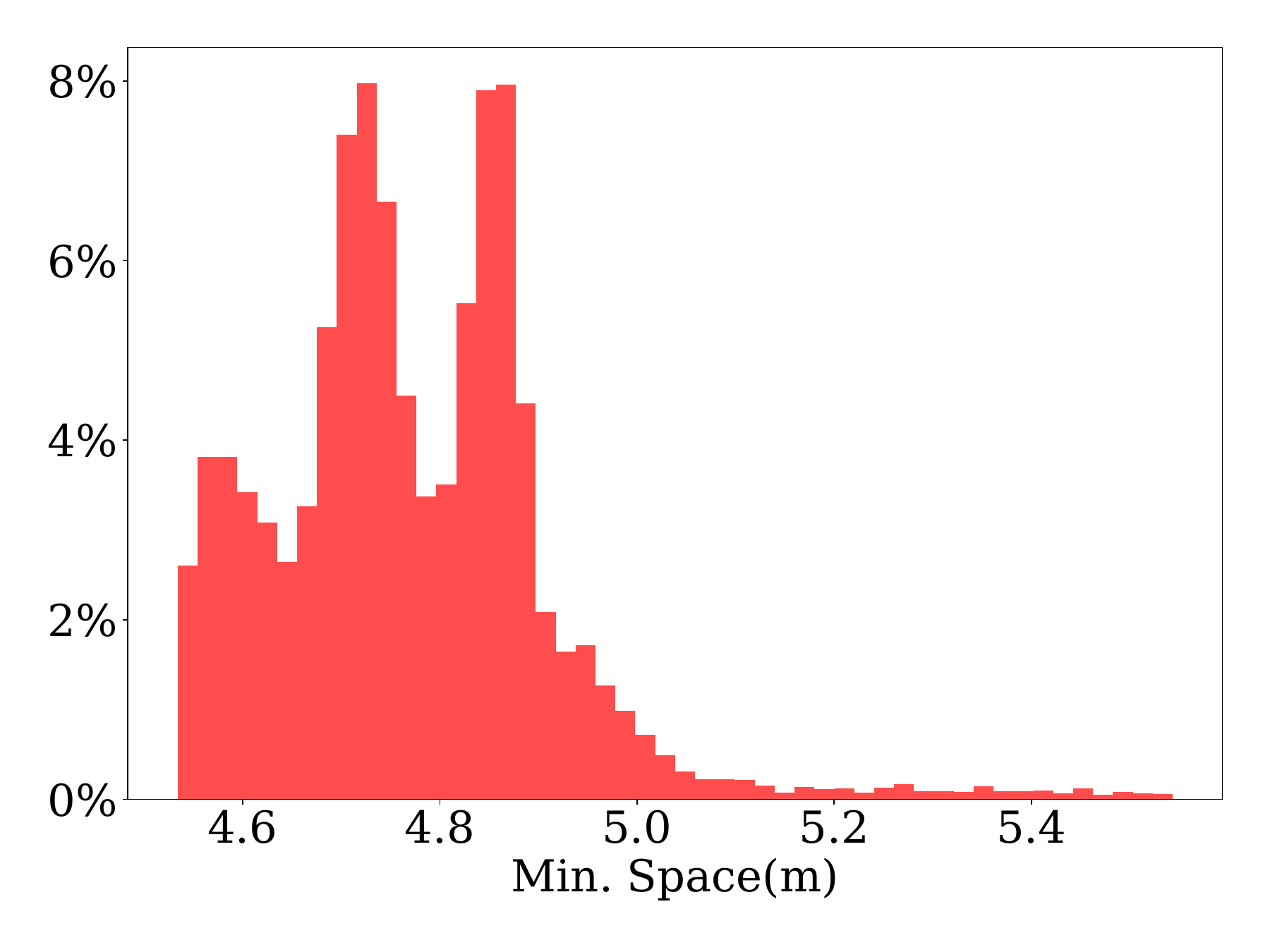}
    \end{subfigure}
    \begin{subfigure}[b]{0.18\linewidth}
        \centering
        \includegraphics[width=\linewidth]{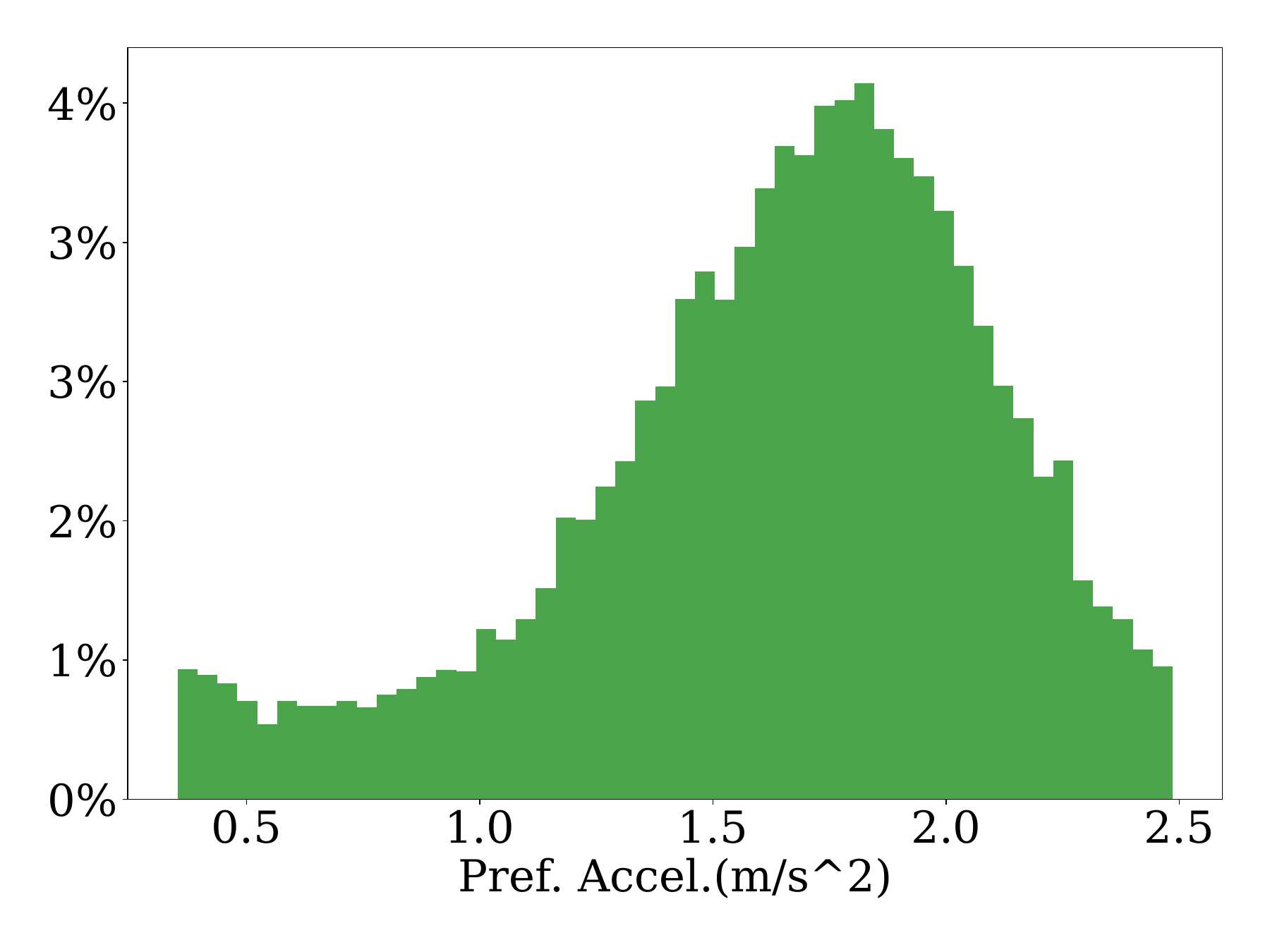}
    \end{subfigure}
    \begin{subfigure}[b]{0.18\linewidth}
        \centering
        \includegraphics[width=\linewidth]{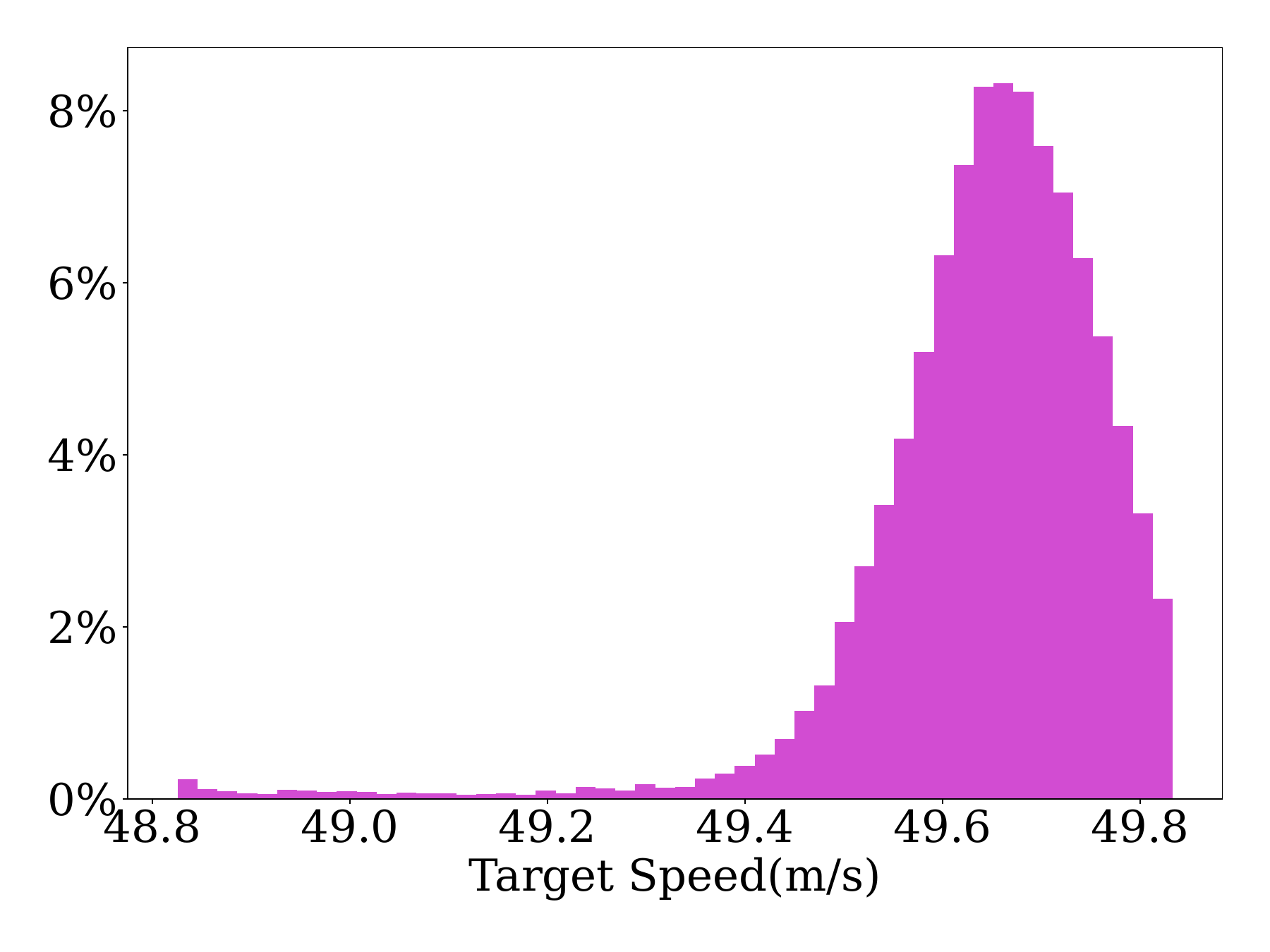}
    \end{subfigure}
    \begin{subfigure}[b]{0.18\linewidth}
        \centering
        \includegraphics[width=\linewidth]{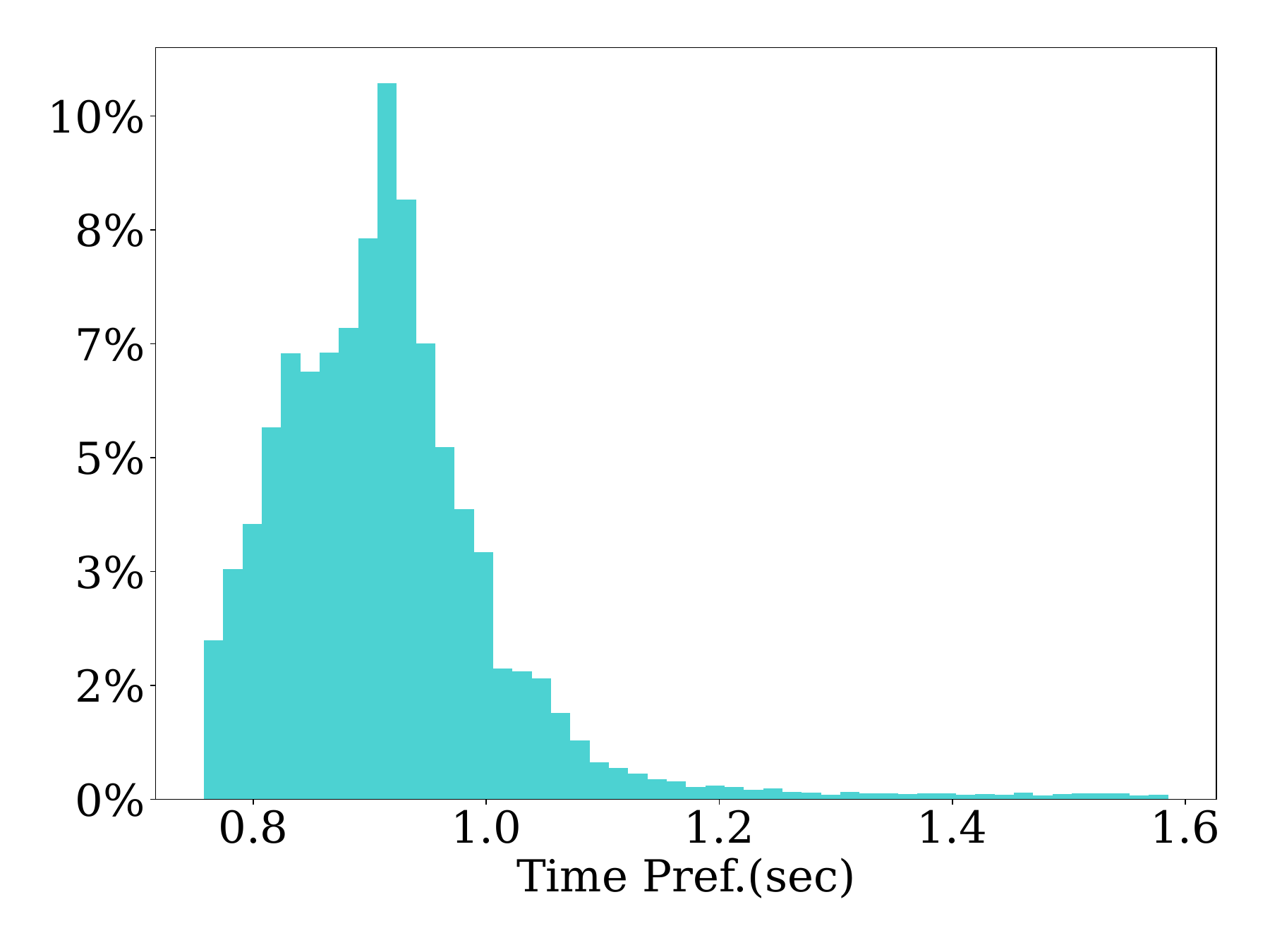}
    \end{subfigure}
    \caption{\textbf{Distribution of optimized IDM parameters in trajectory reconstruction task.} Note that the distribution of each parameter adheres well to the general, real-world scenarios.}
    \label{fig:traj-recon-idm}
    \vspace{-1em}
\end{figure*}

\subsection{Trajectory Filtering \& Reconstruction}

Since the trajectory filtering and reconstruction tasks differ only in data point density, we apply the same solution to both. We assume that each trajectory is represented by a sequence of timestamps and corresponding vehicle positions in a 1-dimensional space. For the $i$-th trajectory, let $\mathbb{T}^{i}_{j}$ (sec) be the $j$-th timestamp and $\mathbb{P}^{i}_{j}$ (m) the corresponding vehicle position. Our goal is to generate dense trajectories that interpolate these points while adhering to physical laws. Simple linear interpolation often leads to sudden changes in estimated speed and acceleration, which are physically infeasible (Figure~\ref{fig:traj-profile}).

Therefore, we use our traffic simulator to reconstruct physically consistent trajectories that adhere to the given data as much as possible. To be specific, if we denote the temporal length of an $i$-th trajectory as $\mathbb{T}^{i}_{l}$, we simulate the motion of the ego vehicle from time $0$ to $\mathbb{T}^{i}_{l}$, using $\delta t$ as $\Delta t$ in Eq.~\ref{eq:euler}. Then, for each time stamp $\mathbb{T}^{i}_{j}$ in the input, we can find the nearest time step $k^{i}_{j} \cdot \delta t$, where $k^{i}_{j}$ is a positive integer, and compute the reconstruction loss for the $i$-th trajectory based on mean absolute error (MAE):
\begin{equation}
    L^{i}_{recon} = \sum_{j=1}^{l} |\mathbb{P}^{i}_{j} - \mathbf{P}[k^{i}_{j}]|,
\end{equation}

\noindent
where $\mathbf{P}$ is an array that stores the position of the vehicle during simulation. As our simulator supports large-scale simulation, we simulate all trajectories at once and compute the final loss $L = \sum_{i=1}^{N}L^{i}_{recon}$.

Once the reconstruction loss is computed, we minimize it using gradient-based optimization techniques.
We optimize five IDM parameters $(a_{max}, a_{pref}, T_{pref}, s_{min}, v_{targ})$, which represent driver behavior and remain fixed for each trajectory. These parameters are initialized to 10, 2, 1, 5, and 50, respectively, with constraints of $[5, 10], [0.1, 5], [0.1, 5], [1, 10], [20, 60]$. We also optimize the lists of $\Delta p_k$ and $\Delta v_k$ for each simulation step $k$, initializing them to 10 and 0. Lastly, $a_{min}$ is set to $-10$ and remains fixed during optimization.



\subsection{Trajectory Prediction}

In trajectory prediction, the goal is to forecast a future trajectory based on a historical trajectory and scene context. Recent open-loop deep learning methods use neural networks to identify patterns from real-world trajectory data and predict human-like outcomes~\cite{shi2024_mtrv3, sun2024controlmtr, ettinger2024scaling, mppp_multipath_varadarajan_2022, nayakanti_wayformer, seff2023motionlm}. However, common failure modes early in training involve scenarios where vehicles collide with the leading vehicle or go beyond road boundaries. Additionally, deep neural network approaches for trajectory prediction are known to be data inefficient, despite achieving SOTA results~\cite{bharilya2024machine}.

We use our differentiable traffic simulation to provide a {\em learning-free} performance baseline using differentiable ODE modeling and inferred road context. First, we infer traffic states from vehicle trajectories and lane shapes. For our experiments, we use the Waymo Open Motion Dataset (WOMD)~\cite{womd_Ettinger_2021_ICCV}. Agents are assigned to lanes based on proximity to lane polyline features, and both lane polylines and agent states are projected into 1-dimensional representations for our simulator. We then fit IDM parameters using the historical trajectory, similar to the optimization process in the trajectory reconstruction task. In deep learning approaches, this information would be encoded into context embeddings. Finally, we simulate trajectories in the 1-dimensional space using the fitted IDM parameters and project them back onto the original lane polylines.

We hypothesize that by using this approach, the ``predicted" trajectory will closely align with real-world trajectories, without requiring deep neural networks for learning. If successful, these ODE-based rollouts could provide a strong baseline and lower bound on performance metrics for more complex, learning-based methods.

This task is not feasible with existing simulators. While some simulators use ODE-based traffic state construction, their lack of differentiability or parallelization makes them impractical for large-scale trajectory prediction. Conversely, large-scale parallelized simulators do not explicitly construct traffic states from road data or use ODEs to model agents, making them heavily dependent on learned policies and real-world data logs for future rollouts.

\section{Experimental Results}

In this section, we provide experimental results. First, we present the experimental results about the efficiency of our simulator. Then, we provide results of trajectory filtering and reconstruction tasks together and discuss those of trajectory prediction in the end. Our approach is implemented in Python and PyTorch~\cite{paszke2019pytorch}. All of our experiments were done on a workstation equipped with Intel\textregistered~Xeon\textregistered~W-2255 CPU @ 3.70GHz and a single NVIDIA RTX A5000 GPU. 

\begin{table}[]
\caption{\textbf{Quantitative comparison for trajectory filtering \& reconstruction task.} For each task, we compare the average positional error rate (Pos.), magnitude of estimated acceleration (Acc.), physically implausible trajectory rate (Imp.), and average computation time per trajectory (Time.) between compared methods. For estimated acceleration magnitude ($m/s^2$), we provide mean and standard variation ($\mu/\sigma$).}
\footnotesize
\centering
\begin{tabular}{c|c|cccc|c}
\toprule
Task & Value & Linear & MA & EMA & Spline & Ours \\ 
\midrule
{\multirow{4}{*}{Fil.}} & Pos.$\mathbf{\downarrow}$  & 0.00\% & 0.01\% & 0.02\% & 0.00\% & 0.08\% \\ 
                         & Acc.$\mathbf{\downarrow}$  & $6.4/5.5$ & $0.9/0.9$ & $0.5/0.6$ & $3.5/2.1$ & $0.5/0.5$ \\
                         & Imp.$\mathbf{\downarrow}$  & $100\%$ & $8.18\%$ & $4.61\%$ & $2.00\%$ & $0\%$ \\
                         & Time.$\mathbf{\downarrow}$ & 0.115s & 0.233s & 0.237s & 23.9s & 0.308s \\ 
\midrule
{\multirow{4}{*}{Rec.}}  & Pos.  & 0.00\% & 0.05\% & 0.06\% & 0.01\% & 0.13\% \\ 
                         & Acc.  & 1.8/5.4 & 0.5/0.7 & 0.2/0.4 & 1.6/1.9 & 0.3/1.1 \\ 
                         & Imp.  & $99.84\%$ & $9.83\%$ & $7.24\%$ & $4.12\%$ & $0\%$ \\
                         & Time. & 0.033s & 0.038s & 0.042s & 3.37s & 0.164s \\ 
\bottomrule
\end{tabular}
\label{tab:traj-recon}
\end{table}

\subsection{Computational Efficiency}

To assess the computational efficiency of our simulator, we measured its running time for both the forward and backward passes. The total time was divided by the number of simulation steps, as running time scales proportionally with the number of steps. To evaluate scalability and device impact, we tested different numbers of vehicles using either CPU (with 16 threads) or GPU. As shown in Figure~\ref{fig:cost-comp-device}, our simulator can handle up to 2 million vehicles in real-time on both CPU and GPU. The CPU performs efficiently due to the use of 16 threads. For smaller numbers of vehicles, the CPU is faster since launching GPU jobs introduces overhead. However, the GPU demonstrates better scalability as the number of vehicles goes beyond $20K$.

\subsection{Trajectory Filtering \& Reconstruction}

\subsubsection{Dataset}

For the trajectory filtering task, we used the publicly available Next Generation Simulation (NGSIM) dataset~\cite{alexiadis2004next}, specifically 6,101 trajectories from US Highway 101, recorded every 100 milliseconds, providing dense data for filtering. For trajectory reconstruction, we used a non-public dataset containing 21,750 sparse trajectories from Singapore, recorded with a minimum sampling interval of 1 second. Each trajectory is up to 20 minutes long.

\subsubsection{Baselines}

For comparison, we used two interpolation methods (linear, spline) and two filtering techniques (moving average~\cite{duret2008estimating, ossen2008validity}, exponential moving average~\cite{thiemann2008estimating}) as baselines. For interpolation, we set $\delta t$ to 0.1 and 1.0 for each task, estimating vehicle positions at every $\delta t$ to generate dense trajectories. This matches our method, where $\delta t$ is the simulation time step. For the filtering task, linear interpolation results were identical to the raw data, as the data was recorded \textit{every} 100 milliseconds. For spline interpolation, we fitted a cubic B-spline curve using Scipy's optimization function~\cite{virtanen2020scipy}, setting bounds on the first and second derivatives (speed and acceleration) to $[0, \infty)$ and $[-10, 10]$, respectively, to ensure realistic trajectories as much as possible. For filtering, we applied both methods to the dense trajectories from linear interpolation, using a window size of 9 for moving average (MA) and a smoothing width of 5 for exponential moving average (EMA), following~\cite{ossen2008validity} and~\cite{thiemann2008estimating}. 

\subsubsection{Optimization Details}

In our optimization process, we used the Adam optimizer~\cite{kingma2014adam} with 500 optimization steps for both tasks. During optimization, we linearly reduced the learning rate from an initial value of 0.1 to 0.01. The initial position $p(0)$ and speed $v(0)$ for each trajectory were set to 0 and $(\Delta \mathbb{P}) / \Delta t$, where $\Delta \mathbb{P}$ is the distance between the first two data points.

\begin{figure}
    \centering
    \includegraphics[width=\linewidth]{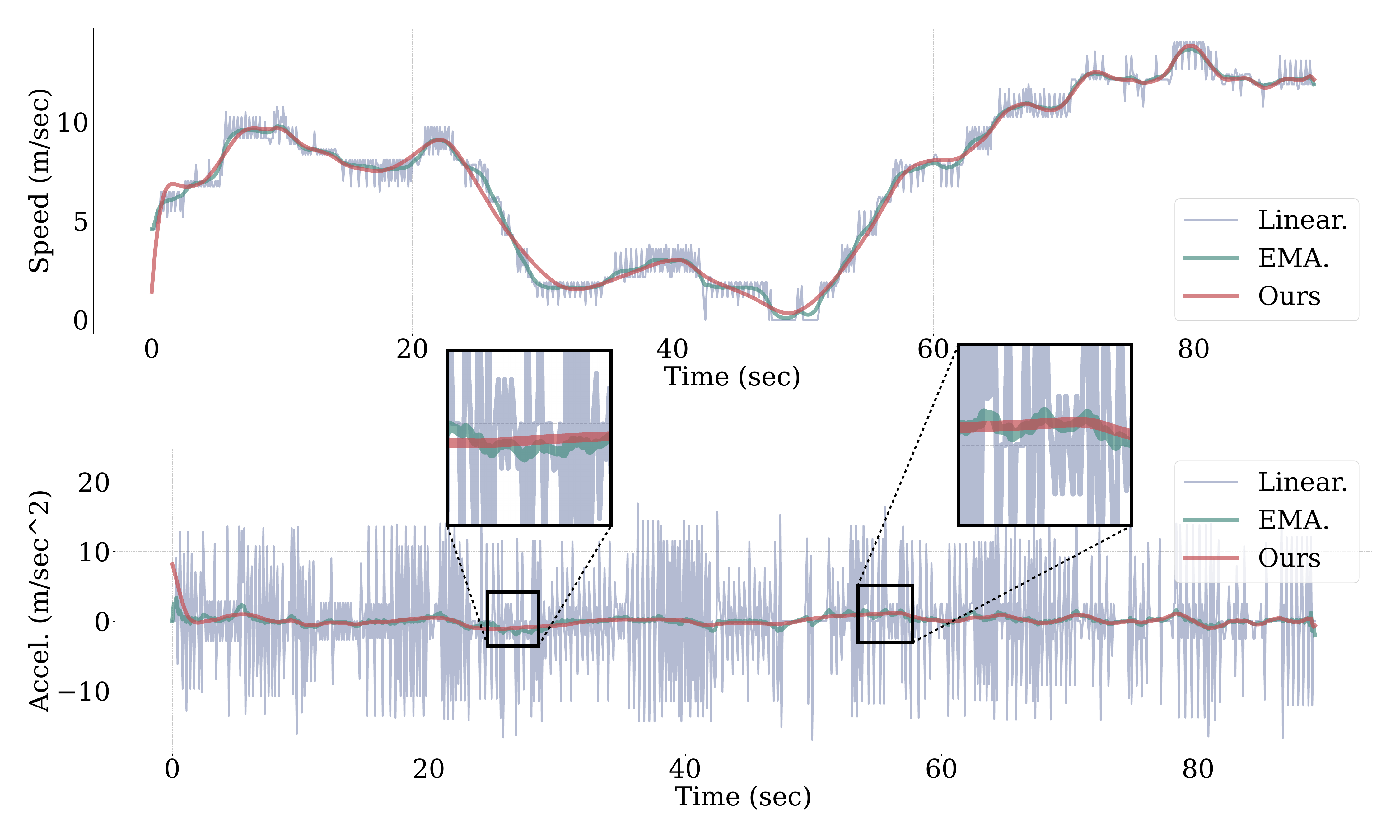}
    \caption{\textbf{Qualitative comparison for trajectory filtering task.} For comparison, we use the speed (up) and acceleration (down) profiles for a single trajectory in the NGSIM dataset. Estimated acceleration from our method exhibits a more stable pattern than the baseline methods.}
    \vspace{-1em}
    \label{fig:traj-profile}
\end{figure}

\subsubsection{Quantitative Comparison}

Table~\ref{tab:traj-recon} presents quantitative results based on four criteria:
\begin{enumerate}
    \item \textbf{Average positional error rate (Pos.)}: Represents the precision of the trajectory, calculated by finding the nearest time step in the estimated dense trajectory for each data point, measuring the distance, dividing by the total trajectory length, and averaging across all points.
    \item \textbf{Statistics about estimated accelerations (Acc.)}: Mean and standard deviation of absolute accelerations. Lower values usually indicate more stable trajectories.
    \item \textbf{Ratio of physically implausible trajectories (Imp.)}: Ratio of trajectories with any step where absolute acceleration exceeds 10, as most real-world absolute accelerations stay below 3.
    \item \textbf{Running Time (Time.)}: Running time per trajectory.
\end{enumerate}

As we can see in Table~\ref{tab:traj-recon}, linear and spline interpolation achieved the best results in positional error, but most of their trajectories were physically invalid. MA and EMA also performed well in positional error and gave impressive results on the second criterion. However, they still produced physically invalid trajectories, and they are not negligible---one of the trajectories exhibited an acceleration of 66, which is physically impossible. In contrast, \textbf{\textit{trajectories from our method were all physically valid, and showed stable acceleration pattern, with only a slight loss in positional error}}. In terms of computational cost, our method was up to 5 times slower than simple linear interpolation \textbf{\textit{but processed each trajectory in under 0.4 seconds with parallelization}}.

\begin{figure}
    \centering
    \includegraphics[width=0.95\linewidth]{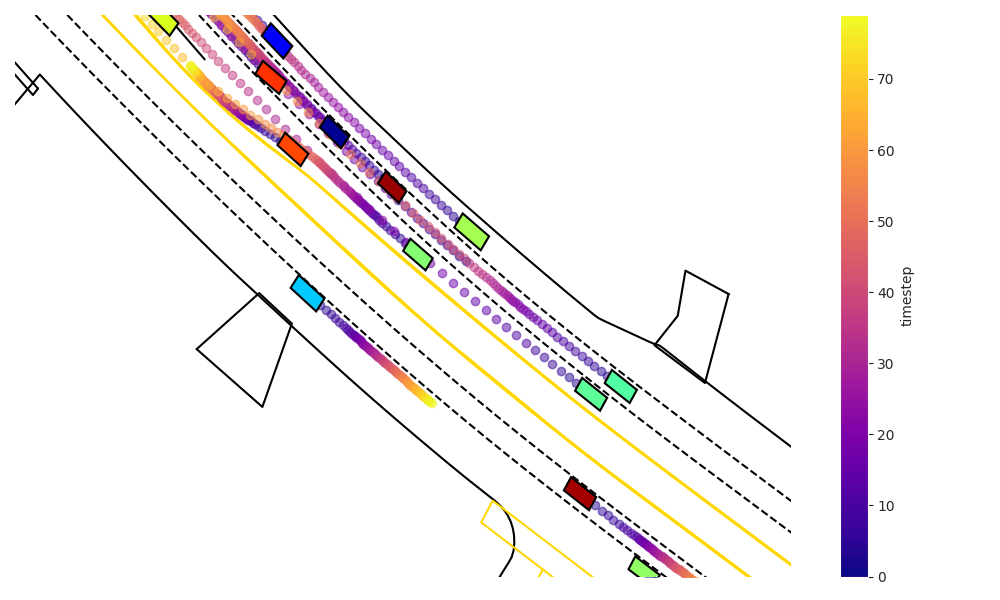}
    \caption{\textbf{Training-free future rollouts on the Waymo Open Motion Dataset (WOMD).} We show qualitative results in trajectory forecasting with our differentiable traffic simulator. In the trajectory forecasting task, we infer the traffic state based on road graph and agent observations, including agent membership and order within lanes. Then, we fit IDM parameters for each vehicle to its 1-second historical trajectory and roll out trajectories via IDM for all agents according to their current lane. As shown above, agent vehicles maintain realistic trajectories with no deep learning or ground truth.}
    \label{fig:waymo-qualitative}
    \vspace{-1em}
\end{figure}

\subsubsection{Qualitative Comparison}

To qualitatively assess the optimized trajectories, we plot the speed and acceleration profile of a trajectory from the NGSIM dataset in Figure~\ref{fig:traj-profile}, comparing our results with linear interpolation and EMA. Linear interpolation simply mirrors the raw data (possibly with noise), so the resulting speed and acceleration profiles contain significant physically invalid noise. After applying the EMA filter, high-frequency noise is reduced, and both profiles become smoother -- however, it is still not guaranteed to be physically valid. Our method produces even smoother results, and they are guaranteed to be physically valid.

\subsubsection{IDM Parameters}

As mentioned earlier, we optimize five IDM parameters during trajectory optimization. To assess the impact of optimizing these variables, we reran the trajectory reconstruction task with fixed IDM parameters. As expected, this produced worse results than our original approach: the positional error rate increased to 0.22\%, and the estimated acceleration became $(0.5/1.7)$. Optimizing the IDM parameters allows for better results by providing more flexibility. Figure~\ref{fig:traj-recon-idm} shows the distribution of the optimized IDM parameters, which closely aligns with real-world observations. This demonstrates that, in addition to producing better results, \textbf{\textit{our method can infer driver behavior during optimization}}, offering an advantage over baseline methods.

\subsection{A Training-Free Trajectory Forecasting Baseline}

In trajectory forecasting experiments, we use our differentiable simulator to fit IDM parameters and perform training-free trajectory rollouts on the Waymo Open Motion Dataset (WOMD)~\cite{womd_Ettinger_2021_ICCV} validation set, which contains 15.5K 20-second segments split into overlapping 9-second samples. Each sample represents a traffic scenario centered on an autonomous vehicle. Using the Adam optimizer, we fit IDM parameters to 1-second trajectory histories sampled at 10Hz over 10 timesteps.
To compute the traffic state as positions along 1-dimensional lanes, we project each vehicle's historical trajectory onto lane centers, where only vehicles within a certain distance threshold of lane centers are considered for future prediction.
After solving for IDM parameters with gradient-based optimization, we initialize the traffic state to the starting state and roll out 8-second futures at 10 Hz, for a total of 80 frames.

For each vehicle active in traffic lanes, we evaluate their performance with WOMD benchmark metrics. 
That is, we show results in mean average precision (mAP), minimum average displacement error (minADE), minimum final displacement error (minFDE), and miss rate (MissRate):
\begin{enumerate}
    \item \textbf{mAP}: Mean average precision across all trajectory types and agent classes, where trajectory types include straight, straight-left, straight-right, left, right, left u-turn, right u-turn, and stationary.
    \item \textbf{minADE}: L2 norm between closest prediction and ground truth.
    \item \textbf{minFDE}: L2 norm between closest final position and ground truth final position.
    \item \textbf{Miss Rate}: \% of agent predictions outside of a threshold from ground truth.
\end{enumerate}

Closer details on metrics can be found in Ettinger et al.'s work introducing the benchmark~\cite{womd_Ettinger_2021_ICCV}. 

\begin{table}[t!]
 \caption{{\bf Waymo Open Motion Dataset validation set metric comparisons between SOTA models and training-free rollouts generated by our differentiable simulator.} We show metric performance results generated from our simulator versus current SOTA models for trajectory prediction. Without any neural networks or training, future predictions using our simulator can attain 0.3541 mAP on validation---while this result does not outperform SOTA models, this performance is attained with {\em no training}, where IDM parameters are only fitted on {\bf 1-second of data}. Missing values are not reported by the respective original work.}
\label{tb:womd_comparison}
  \centering
  \scalebox{0.8}{
  \begin{tabular}{l|rrrrr}
    \toprule
    Method &  mAP$\uparrow$ &  minADE$\downarrow$ &  minFDE$\downarrow$ &  MissRate$\downarrow$ & Params\\
    \midrule
Multipath++~\cite{mppp_multipath_varadarajan_2022} & 0.393 & 0.978 & 2.305 & 0.440 & - \\
MTR~\cite{shi2022motion}& 0.416 & 0.605 & 1.225 & 0.137 & 66M \\
ControlMTR~\cite{sun2024controlmtr} & 0.422 & 0.590 & 1.201 & 0.132 & 66M \\
Wayformer\_Ens~\cite{ettinger2024scaling} & 0.430 & 0.530 & - & - & 60M \\
MTR++~\cite{shi2023mtr} & 0.435 & 0.591 & 1.199 & 0.130 & 86M \\
EDA~\cite{lin2024eda} & 0.435 & 0.571 & 1.173 & 0.119 & - \\
MTRv3\_Ens~\cite{shi2024_mtrv3} & 0.488 & 0.554 & 1.104 & 0.110 & - \\
\midrule
Ours & 0.354 & 7.862 & 15.668 & 0.586 & \textbf{5} \\
    \bottomrule
  \end{tabular}
  }
\end{table}

We show quantitative results on WOMD validation set in Table~\ref{tb:womd_comparison}, and compare to recent SOTA models in trajectory forecasting~\cite{sun2024controlmtr, ettinger2024scaling, lin2024eda, shi2024_mtrv3, shi2022motion, shi2023mtr}. 
In this result, we consider all benchmark trajectories and agent classes in the validation set, including that of cyclists and pedestrians in addition to vehicles.

Expectedly, results with our traffic simulator does not outperform SOTA methods, and should not be directly compared against models trained on ground truth annotations.
However, \textbf{\textit{we find that our simulator achieves comparable mAP performance (0.354)}} and acceptable distance metric performance (7.862 minADE), \textbf{\textit{given no training on a large amount of pre-annotated data, where the next best mAP performance is Multipath++~\cite{mppp_multipath_varadarajan_2022} at 0.393}}.
We point out that minADE and minFDE values are high perhaps due to poor performance on samples involving non-straight driving. 
In this experiment, vehicles merely advance forward along the current road centers. mAP values may be competitive purely due to overrepresentation of straight driving in benchmark datasets. 
Still, our method provides an interesting blind driving benchmark for trajectory forecasting work, which does not involve any deep learning whatsoever.
We visualize an example rollout on a WOMD validation sample in Figure~\ref{fig:waymo-qualitative}, where we show qualitatively that simulated rollouts are realistic and follow road structures.

\section{Conclusion}


In this paper, we introduced a parallelized differentiable traffic simulator based on IDM, capable of simulating up to 2 million vehicles in real-time on parallel CPU or GPU implementations. To address physically implausible results, we modified the original IDM and demonstrated the simulator's effectiveness in solving trajectory optimization problems. For trajectory filtering and reconstruction, our method reliably estimated physically plausible trajectories. Additionally, we showed that differentiable traffic simulation serves as a strong training-free baseline for trajectory forecasting and paves the way for integrating traffic simulation with deep learning for robust, traffic-aware techniques.

A limitation of our work is its focus on IDM-based vehicle modeling, which could be extended to include ODE-based pedestrian and cyclist behavior for more robust agent representations. Future work could also explore more complex loss functions for improved trajectory accuracy and support for complex road networks. Nonetheless, our simulator can serve as a core computational layer for integrating with traffic simulators that handle such systems.


\vspace*{0.5em}
\noindent
{\bf Acknowledgment:  }  This research is supported in part by Barry Mersky and Capital One E-Nnovate Endowed Professors, the ARO DURIP Grant, and the HAYSTAC Program.









\bibliographystyle{IEEEtran}
\bibliography{ref}

\onecolumn
\section*{APPENDIX}

\subsection{Feature Comparison with Existing Vehicle Simulators}
\begin{table}[h!]
    \centering
    \begin{tabular}{c|cccc}  
        \toprule
        Simulator & Parallelized & GPU Acceleration & ODE-Differentiable & Log Replay \\
        \midrule
        HighwayEnv~\cite{highway-env} & \xmark & \xmark & \xmark & \xmark \\
        SUMO~\cite{SUMO2018} & \xmark & \xmark & \xmark & \xmark \\
        Waymax~\cite{gulino2024waymax} & \cmark & \cmark & \xmark & \cmark \\
        MOSS~\cite{zhang2024moss} & \cmark & \cmark & \xmark & \xmark \\
        GPUDrive~\cite{gpudrive_Kazemkhani_Pandya_Cornelisse_Shacklett_Vinitsky_2024} & \cmark & \cmark & \xmark & \cmark \\
        Ours & \cmark & \cmark & \cmark & \cmark \\
        \bottomrule
    \end{tabular}
    \caption{Feature comparisons between existing popular traffic simulators and ours. Note that this comparison is for vehicle simulation only.}
    \label{tab:appendix_simulator_comparison}
\end{table}

\twocolumn
\balance
\end{document}